%% file: root.tex
\documentclass[letterpaper, 10 pt, conference]{ieeeconf}  

\IEEEoverridecommandlockouts                              

\usepackage{amssymb}
\usepackage{amsmath}
\usepackage{graphicx}
\usepackage{xcolor}
\usepackage{listings}
\usepackage{multirow}
\usepackage{color}
\usepackage{pifont}
\usepackage{array} 
\usepackage{makecell}
\usepackage{booktabs}
\title{\LARGE \bf
COMMET: A System for Human-Induced Conflicts in\\ Mobile Manipulation of Everyday Tasks
}

\author{Dongping Li$^{1,2*}$, Shaoting Peng$^{1*}$, John Pohovey$^{1}$, Katherine Rose Driggs-Campbell$^{1}$
\thanks{* Equal Contribution}
\thanks{$^{1}$ D. Li is a visiting student of University of Illinois Urbana-Champaign. S. Peng, J. Pohovey and  K. Driggs-Campbell are with the Department of Electrical
and Computer Engineering at University of Illinois Urbana-Champaign. emails: {\tt\small \{dongping, peng33, jpohov2, krdc\}@illinois.edu}}
\thanks{$^{2}$ D. Li is also with ZJU-UIUC Institute, Zhejiang University. emails: {\tt\small dongping.23@intl.zju.edu.cn}}
}

\begin{document}

\maketitle
\thispagestyle{empty}
\pagestyle{empty}

\input{tex/0_abs}
\input{tex/1_intro}
\input{tex/2_related_work}

\input{tex/3_system}

\input{tex/4_exp}

\input{tex/5_conclusion}



\bibliographystyle{ieeetr}
\bibliography{ref}

\newpage
\input{tex/6_appendix}

\end{document}

%% file: tex/0_abs.tex
\begin{abstract}
Continuous advancements in robotics and AI are driving the integration of robots from industry into everyday environments. However, dynamic and unpredictable human activities in daily lives would directly or indirectly conflict with robot actions. Besides, due to the social attributes of such human-induced conflicts, solutions are not always unique and depend highly on the user's personal preferences. To address these challenges and facilitate the development of household robots, we propose \textbf{COMMET}, a system for human-induced \textbf{CO}nflicts in \textbf{M}obile \textbf{M}anipulation of \textbf{E}veryday \textbf{T}asks. COMMET employs a hybrid detection approach, which begins with multi-modal retrieval and escalates to fine-tuned model inference for low-confidence cases. Based on collected user preferred options and settings, GPT-4o will be used to summarize user preferences from relevant cases. In preliminary studies, our detection module shows better accuracy and latency compared with GPT models. To facilitate future research, we also design a user-friendly interface for user data collection and demonstrate an effective workflow for real-world deployments.
\end{abstract}

%% file: tex/1_intro.tex
\section{Introduction}
\label{sec:intro}
Intelligent household robots have been a long-standing goal in robotics. In recent years, advancements in Large Language Models (LLMs)~\cite{achiam2023gpt,guo2025deepseek} and embodied intelligence~\cite{ahn2022can,driess2023palm} have greatly accelerated progress toward this goal, endowing robots with powerful abilities in perception, planning, navigation and manipulation~\cite{song2023llm,kim2024openvla,brohan2023rt,chen2023predicting,yokoyama2024vlfm,liu2024ok,fu2024mobile}. Besides, considering the social attribute of household robots, various Human-Robot Interaction (HRI) studies~\cite{liu2024dragon,song2024vlm,peng2025towards,wang2024apricot,wu2023tidybot} aim to facilitate robots better integrate into people's daily lives. Meanwhile, works about adaptive methods and failure detection~\cite{sinha2024real,xiong2024adaptive,liu2024multi,li2025emmoe} focus on enabling robots to handle real-world disturbances. However, as opposed to more controllable environments such as laboratories or simulations, \textbf{dynamic human activities in the real world will induce more uncontrollable factors, which could cause various conflicts with the robot's actions}. Indirect conflicts like changes in environmental states or object locations; Direct conflicts like obstructing or interrupting the robot's planning and executions. These conflicts will influence robots at both low levels and high levels. Although some works have explored robots in dynamic environments~\cite{sinha2024real,liu2024dynamem,yan2024dynamic}, the discussion of the impact of human activities remains limited. Additionally, \textbf{since multiple persons may simultaneously exist in a household environment, the relationships between each individual and the robot would be more complex}. Humans can either be the subjects who instruct and collaborate with robots, or they can be the disturbances that are unrelated to the robot's tasks, but current HRI researches mainly focus on the former cases and overlook interactions when humans serve as disturbances.

\begin{figure}[t]
  \centering
  \includegraphics[width=\linewidth]{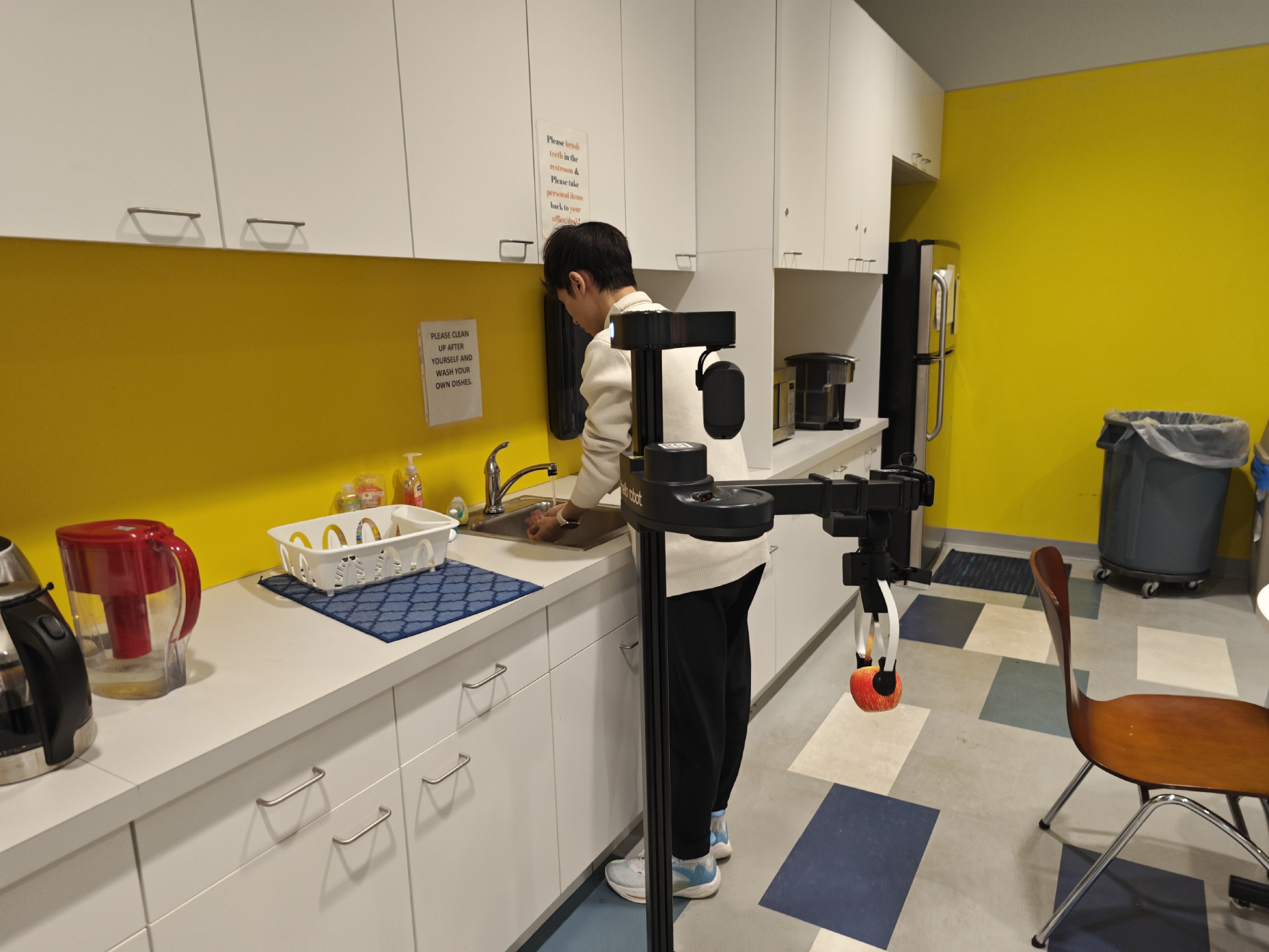}
  \caption{\textbf{An illustration for human-induced conflicts.} The robot is ordered to put the apple into the sink, but another person is washing hands now.}
  \label{fig:demo}
\end{figure}

Moreover, the \textbf{solution to the human-induced conflict is not unique and it largely depends on the user's preferences and the current task context}. For a conflict shown in Fig.~\ref{fig:demo}, if the robot wants to finish user tasks, feasible solutions would include \textit{stop and wait}, \textit{go to another sink}, \textit{ask the user for help}, or \textit{tell the person to leave}, and the optimal option depends on the user's demand. If the user is in a hurry, \textit{tell the person to leave} would be the best choice. Therefore, when robots are trying to solve such conflicts, they should also take user's preferences into account.

To address these challenges, we propose \emph{COMMET}, a system designed to detect and resolve various human-induced conflicts. \emph{COMMET} contains both offline and online stages. In the offline stage, we first collect static scenario and dynamic trajectory data that capture normal cases and potential conflict types in real life, then construct retrieval buffers and model training sets. Next, users need to annotate selected data by providing their preferred solution option and an emergency level which reflects the user's concern level of the current conflict.
In the online stage, \emph{COMMET} starts with the real-time conflict detection, which compares the current robot states and sensor inputs with a speech retrieval buffer and a multi-modal task attribute retrieval buffer. The retrieval output involves the potential conflict type and a confidence score. When the score falls below a pre-set threshold, a fine-tuned LLM will be called for further inference. Once a conflict is detected, \emph{COMMET} will extract corresponding user data based on the detected conflict type, then use GPT-4o~\cite{achiam2023gpt} to summarize the user's preferences and provide the final prediction. In preliminary studies, we analyze the accuracy and latency of our detection module and compare it against GPT~\cite{achiam2023gpt} to show its efficiency and effectiveness. We also design a user-friendly interface to collect user preference data and investigate user ratings of preference prediction. Finally, we discuss how to deploy \emph{COMMET} in the real world and show a potential running workflow.

In summary, our paper makes the following contributions:
\begin{itemize}
\item \textbf{Multi-stage Conflict Detection:} \emph{COMMET} will first conduct real-time speech retrieval and multi-modal task attribute retrieval, and use a fine-tuned model if the retrieval confidence falls below a threshold, the specific conflict type will also be determined simultaneously.
\item \textbf{Fine-grained User Preference Prediction:} 
Except solution options, an emergency level is set to reflect the user's concern level. The input user cases will also vary from the conflict type, then a LLM is asked to summarize user's preferences and provide final predictions.
\item \textbf{System Performance and Feasibility Analysis:} We design a user-friendly interface to facilitate future user studies. We also compare our system's detection performance with GPT-4o, and discuss the possibility and an effective workflow for real-world deployment.
\end{itemize}


%% file: tex/2_related_work.tex
\section{Related Work}
\label{sec:related_work}

\subsection{Intelligent Household Robots}
Intelligent household robots encompass a wide range of robotic topics. Typical imitation learning and reinforcement learning focus on specific actions or primitives~\cite{fu2024mobile,brohan2022rt,shafiullah2023bringing,fang2023anygrasp}. In recent years, LLMs~\cite{achiam2023gpt,guo2025deepseek} have shown powerful generalization and commonsense reasoning abilities and are integrated into more and more robotic works. In zero-shot settings, LLMs can accomplish a wide range of tasks, including planning, manipulation and navigation~\cite{song2023llm,chang2023goat,yokoyama2024vlfm,nagarajah2016hi,huang2024rekep,huang2023voxposer}. Visual Language Action (VLA) models~\cite{kim2024openvla,brohan2023rt,zhang2024navid,sridhar2024nomad} enable models to directly output low-level controls. There are many works that emphasize mobile manipulation in the open world~\cite{liu2024ok,liu2024dynamem,yenamandra2023homerobot,li2025emmoe}. However, dynamic human activities may interfere the robot's actions in various ways. Although some works have explored robots working in dynamic environments~\cite{sinha2024real,xiong2024adaptive,liu2023reflect,liu2024dynamem,yan2024dynamic,yang2024harmonic,zhi2024closed}, the discussion of human-induced conflicts remains limited. Therefore, we categorize common types of daily conflicts induced by humans and propose a multi-stage detection approach to identify these conflicts.

\subsection{Human-Robot Interactions}
Since household robots are unavoidable to contact with humans, many studies are dedicated to enabling robots to seamlessly integrate into people's daily lives. Human-robot collaboration~\cite{wangmosaic,grannen2024vocalsandboxcontinuallearning,shi2024yellrobotimprovingonthefly,ye2023improved} allows people to guide robots or complete tasks together with robots. Human intent estimation~\cite{hoffman2023inferring,belardinelli2024gaze,mascaro2023intention,an2023cooperative} enables robots to anticipate human actions in advance and adjust their actions, thus avoiding potential conflicts or improving people's comfort. Assistive robots~\cite{liu2024dragon,ringwald2023should,irfan2023personalised,ackerman2023robot} are designed to help the elderly or disabled. Human preference learning~\cite{peng2025towards,wang2024apricot,wu2023tidybot,azar2024general,hejna2024contrastive} aims to make the robot's actions better aligned with users' specific needs. However, human-robot relationships in real home environments are more complex. Current research primarily focuses on humans as subjects, overlooking that human interactions and actions can also lead to conflicts. Our work aims to explore a broader range of human-robot relationships and address conflicts caused by human behaviors while taking user preferences into account.

%% file: tex/3_system.tex
\begin{figure*}[t]
  \centering
  \includegraphics[width=\textwidth]{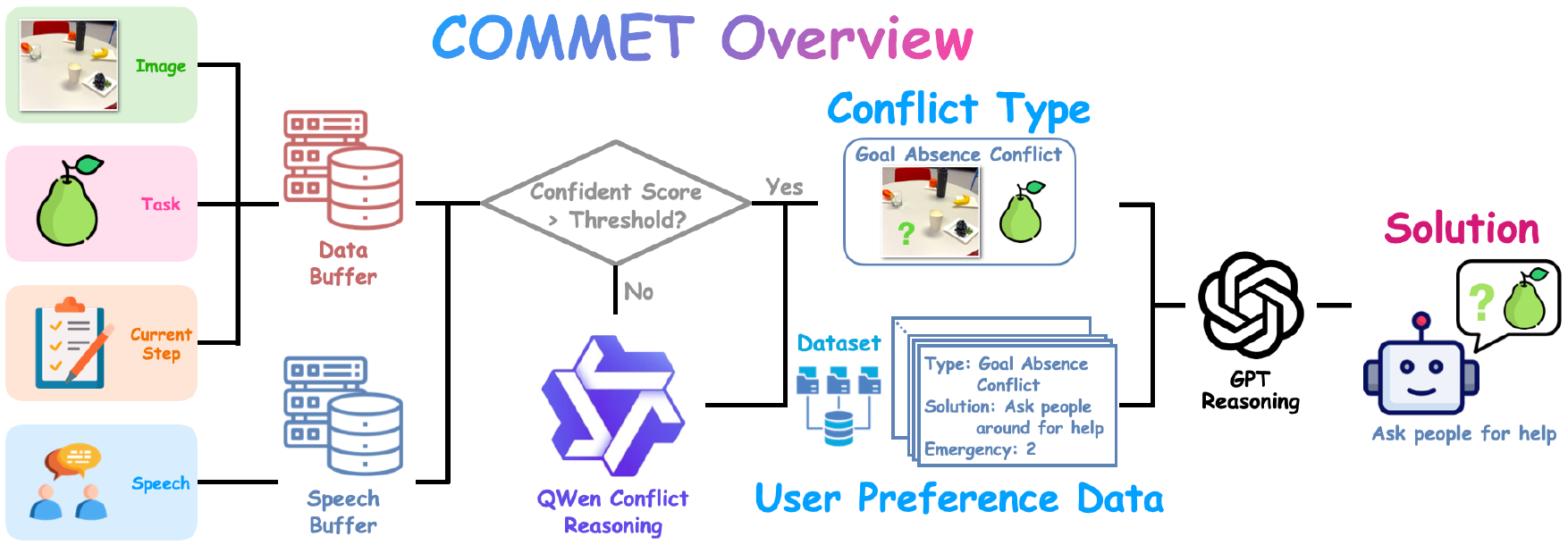}
  \caption{\textbf{System Overview}}
  \label{fig:system}
\end{figure*}

\section{COMMET}
\label{sec:system}

\subsection{Problem Statement}
Our work first makes following assumptions: 1) all conflicts are raised by humans, rather than the robot's planning or execution abilities. 2) all available options are feasible to solve the conflicts. Then the robot needs to detect indirect and direct conflicts brought by humans during the task execution, determine the specific conflict category, and choose a preferred solution based on stored user cases and settings.

\subsection{System Overview}
As is shown in Fig.\ref{fig:system}. The input of \emph{COMMET} includes four components: the real-time visual observation, the final user task, the current step, and the transcribed background speech. We first build two retrieval buffers based on collected user requests, robot trajectories, and anomalous cases. Then we conduct speech retrieval and task attribute retrieval respectively. If the similarity falls below pre-set thresholds, a fine-tuned multi-modal LLM will be used for further inference. Both methods will return a specific conflict type or normal if no conflict is detected. Then we'll input corresponding user options and emergency levels to GPT-4o, which is prompted to summarize user preferences and select a preferred solution. We'll introduce conflict classification in Section~\ref{sys:classification}, data collection in Section~\ref{sys:data}, conflict detection in Section~\ref{sys:detection}, user preference prediction in Section~\ref{sys:preference}.

\subsection{Conflict Category}
\label{sys:classification}
To conduct more fine-grained detection and facilitate the collection of user preferences, we categorize the most common human-induced conflicts during daily task executions into the following four types: 1) \textbf{Goal Absence Conflict}: the robot is going to operate but the target is missing. 2) \textbf{Human Interaction Conflict}: the robot is executing tasks but another person attempts to command the robot or interact with it. 3) \textbf{Human Occupancy Conflict}: Another person's activities occupy the specific space or object, thus hindering the task execution. 4) \textbf{Object State Conflict}: current goal states can't satisfy the task needs, or the current step can't continue due to object state changes (e.g., the door is closed or the container is full). 

\subsection{Data Collection}
\label{sys:data}
We first design static conflict scenarios in 7 different home or indoor environments and include various attributes such as pets, family members, kitchens, bathrooms, living-rooms and so on. Given that real-world detection is a continuous process, we also simulate a first-person perspective of the robot to collect execution trajectories, which include both normal and anomalous cases. All data will be annotated with the observation, task, current step, speech, and conflict type.

To ensure our system can effectively handle potential disturbances that the robot may encounter in the real world, we process the collected data as follows: 1) The \textbf{detection occasion should be time-sensitive}. As there is always a delay between detecting and actually encountering a conflict, it may disappear if the delay is too long. For instance, the robot detects a distant person blocks its path, but the person may move away before the robot actually arrives, thus no actual conflict occurs. Therefore, we maintain a safety range of approximately one meter, conflicts will only be detected when a person or object enters this zone. 2) To \textbf{resist the interference of noise in the real world}, we incorporate daily human conversations that are unrelated to the robot into the dataset. In such cases, the robot should ignore noise and continue operating its original tasks. Finally, we collect 1759 data samples, consisting of 134 static scenarios and 1625 trajectory data captured from 21 distinct tasks. We provide more details about conflict types and data collection in Appendix~\ref{sup:data}.

\begin{figure*}[h]
\centering
\includegraphics[width=\linewidth]{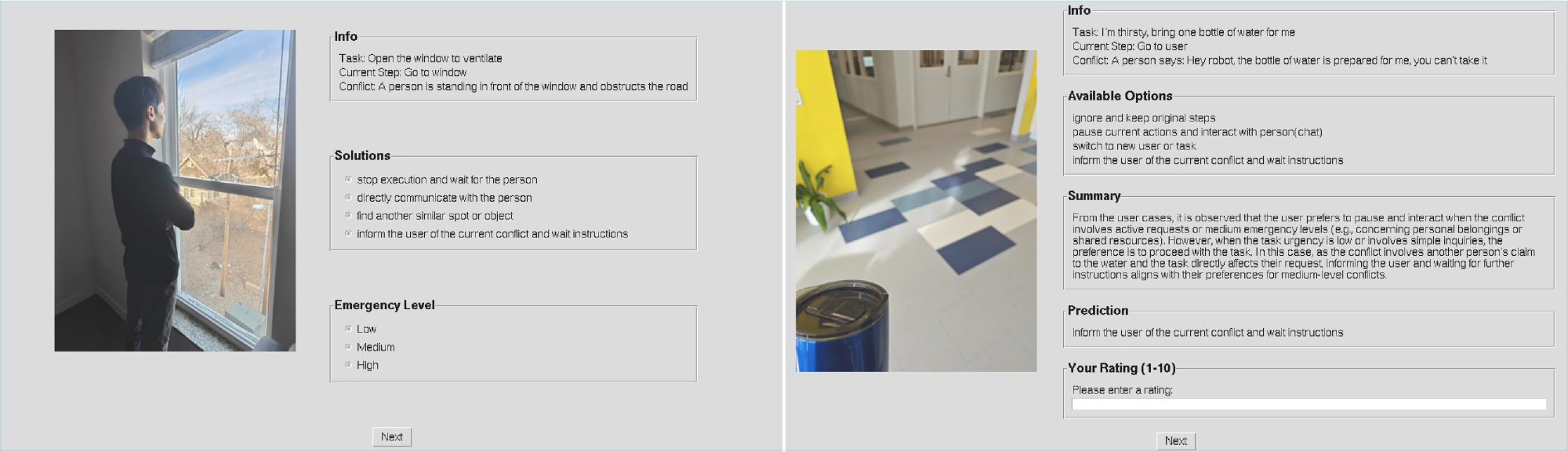}
\caption{\textbf{User Interface.} The left interface collects user preferences, and the right interface collects user final ratings.}
\label{fig:interface}
\end{figure*}

\subsection{Conflict Detection}
\label{sys:detection}
Since real-time detection demands highly on latency, we choose light-weighted embedding models to perform similarity retrieval by default. Specifically, we use gte-large-en-v1.5 (434M)~\cite{zhang2024mgte} to calculate text embeddings and clip (428M)~\cite{radford2021learning} to calculate image embeddings. Moreover, we found that retrieving speech and task attributes (consisting of the user task and the current task step) separately performs better than joint retrieval. Therefore, we design two embedding buffers $B_s$ and $B_m$ based on the data collected in Section~\ref{sys:data}, and conduct two parallel retrieval processes. $B_s$ is a text-only speech buffer, while $B_m$ is a multi-modal embedding buffer that stores both image embeddings and task attribute embeddings.

For speech retrieval, we first convert the detected background audio into text and calculate its embedding $E_s$. Then we calculate the cosine similarity between $E_s$ and each embedding $E_i$ in $B_s$. If the max score $S_s$ surpasses a threshold $\tau_s$, we'll determine that another person is attempting to interact with the robot and directly trigger a \textit{Human Interaction Conflict}. For task attribute retrieval, we calculate both the prompt embedding $E_p$ and image embedding $E_{obs}$ similarity with each $E_i$ in $B_m$, and utilize a weight parameter $w$ to calculate the final score $S_t$. Then corresponding conflict or normal information will be extracted.
\begin{equation}\label{tp} 
S_s = \max_{E_i \in B_s} \left( \cos(E_s, E_i) \right)
\end{equation}
\begin{equation}
S_t = \max_{E_i \in B_m} \left( w \cdot \cos(E_i^{p}, E_p) + (1 - w) \cdot \cos(E_i^{obs}, E_{obs}) \right)
\end{equation}
Although retrieval methods allow us to efficiently leverage existing data with very low latency, manually collected data is limited and often fail to cover all possible conflicts in the real world. It also struggles to generalize to new environments and tasks. These issues are particularly prominent in task attribute retrieval, as the user's demands are various. So we set another threshold $\tau_t$ for task attribute retrieval, if $S_t$ falls below $\tau_t$, we assume the current situation may not be represented in $B_m$, and we will leverage powerful commonsense and reasoning capabilities of LLMs for further detection. In this stage, we use previous data to finetune Qwen 2.5VL-3B and 7B~\cite{bai2025qwen2}, which perform well on multi-modal tasks while maintaining a relatively small parameter size. The model input and output formats are consistent with those in the retrieval stage. We will analyze system latency and hyperparameters $w$, $\tau_s$ and $\tau_t$ in Section~\ref{sec:exp}.

\input{table/solutions}

\begin{figure*}[h]
\centering
\includegraphics[width=\linewidth]{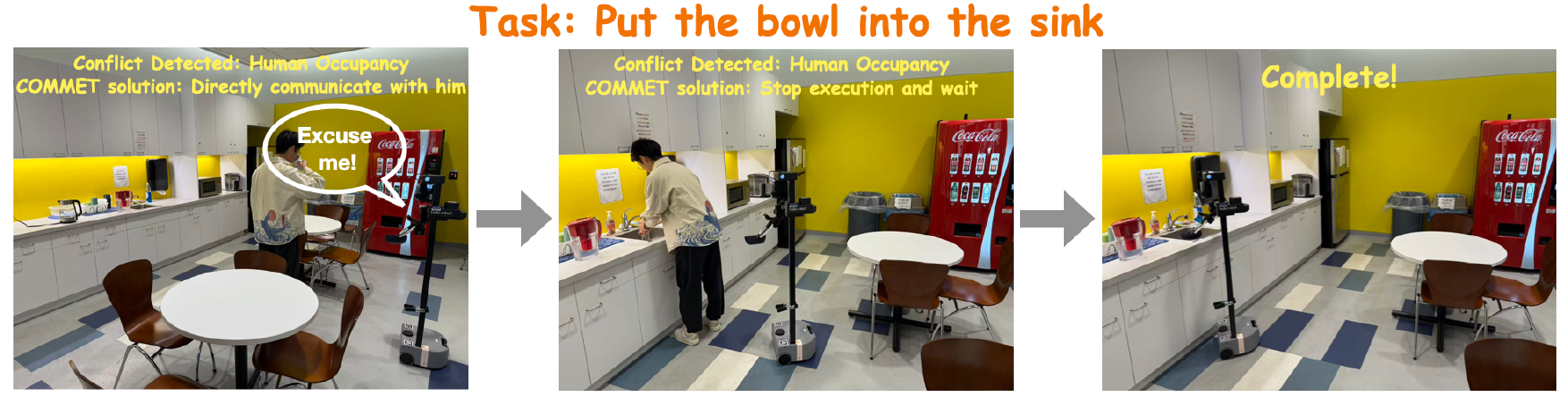}
\caption{\textbf{Real-world Pipeline Example.} Suppose the task is to put the bowl into the sink. On the robot's way to the sink, a human is blocking the way. After identifying the conflict type, \emph{COMMET} reasons based on the user preference data, successfully completing the task.}
\label{fig:real-world}
\end{figure*}

\subsection{User Preference Prediction}
\label{sys:preference}
As is shown in Table~\ref{tab:solutions}, we define several general solutions based on the likelihood of real-world scenarios for each type of conflict. Before the system is running, we'll provide users with 20 conflict scenarios (each type contains 5 cases), they need to select one preferred solution and the emergency level for each scenario. The emergency level has three layers and it reflects the user's concern level and the subjective impact of the current situation: the higher level means the current situation is more urgent to the user. In similar cases, the system should prioritize quickly executing the user’s current task. Likewise, a lower level indicates that users would agree to make certain compromises in time or task performance to resolve conflicts in a more harmonious or easier way. Once a conflict is detected, the current scenario and user data with the same conflict type will be sent together to GPT-4o. It needs to analyze provided user data to generate a summary of user's preferences, then choose the option that best aligns with the user's preferences. More details about detection and prediction can be found in Appendix~\ref{sup:detection} and Appendix~\ref{sup:preference}.



%% file: table/solutions.tex
\begin{table}[t]
\centering
\caption{Available solutions for each type of conflict}
\label{tab:final_solutions}
\renewcommand{\arraystretch}{1.5}
\resizebox{\linewidth}{!}{
    \begin{tabular}{|c|c|}
    \hline
    \textbf{Conflict Type} & \textbf{Solution} \\ \hline
    
    \multirow{4}{*}{\makecell{Goal \\ Absence \\ Conflict}} 
    & Ask people around for help\\ \cline{2-2}
    & Find another similar spot or object \\ \cline{2-2}
    & Re-calculate the path or make a new task plan \\ \cline{2-2}
    & Inform the user and wait for instructions \\ \hline
    
    \multirow{4}{*}{\makecell{Human \\ Occupancy \\ Conflict}}
    & Stop execution and wait for the person \\ \cline{2-2}
    & Directly communicate with the person \\ \cline{2-2}
    & Find another similar spot or object \\ \cline{2-2}
    & Inform the user and wait for instructions \\ \hline
    
    \multirow{4}{*}{\makecell{Object \\ State \\ Conflict}}
    & Ask people around for help\\ \cline{2-2}
    & Find another similar spot or object \\ \cline{2-2}
    & Re-calculate the path or make a new task plan \\ \cline{2-2}
    & Inform the user and wait for instructions \\ \hline
    
    \multirow{4}{*}{\makecell{Human \\ Interaction \\ Conflict}}
    & Ignore and keep original steps \\ \cline{2-2}
    & Pause current actions and interact with person (chat) \\ \cline{2-2}
    & Switch to new user or task \\ \cline{2-2}
    & Inform the user and wait for instructions \\ \hline
    
    \end{tabular}}
\label{tab:solutions}
\end{table}

%% file: tex/4_exp.tex
\input{table/result}

\section{Preliminary Study}
\label{sec:exp}

\subsection{System Hyperparameter and Performance}
For the data collected in Section~\ref{sys:data}, 224 data samples will be used to construct the test set, which includes the complete trajectory data for two tasks and 32 static scenarios. The remaining 1535 data samples will be used to construct the embedding buffer for the retrieval module and the training set for model fine-tuning. 

Next, we determine the values of $w$, $\tau_s$, and $\tau_t$ in sequence. We begin with unified retrieval where the prompt directly incorporates speech text. By varying $w$ from 0 to 1 in increments of 0.01 and observing the corresponding average detection accuracy, we identify the optimal value of $w$ is 0.87. After that, we conduct separate retrieval while $w$ is fixed. Through the same procedure, we obtain the optimal $\tau_s$ of 0.88. Finally, we integrate all modules into \emph{COMMET}, keeping $w$ and $\tau_s$ fixed. We find that the optimal $\tau_t$ is 0.94 when the system is with 3B model and 0.93 with 7B model.

Finally, we measured the average time per detection. Since our data contains both normal and anomalous cases, we also calculate the individual detection accuracy for each case in addition to the total accuracy. Table~\ref{tab:latency} presents performance comparisons across our system, different detection methods, and various detection models. Calculation details can be found in Appendix~\ref{sup:exp}.

\subsection{User Preference Prediction}
we select 10 samples for each conflict type from the static scenario data. As is shown in Fig.~\ref{fig:interface}, users need to provide their preferred solutions and emergency levels for half of these samples. Then each sample in the remaining half will be sent to GPT-4o, along with five samples of the same conflict type from the user data. GPT-4o will summarize the user's selection preferences for the current conflict type and then predict the user's preferred option. After all predictions are completed, users can rate model predictions to provide feedback for future studies.

\subsection{Real-World Deployment}

Since \emph{COMMET} primarily focuses on detecting human-induced conflicts and providing preference-aware solutions, it can be combined with a predefined skill library or an execution system to achieve real-world deployment. An ideal workflow is shown in Fig~\ref{fig:real-world}.

%% file: table/result.tex
\begin{table}[t]
\caption{Performance comparison of models and methods} 
\centering
\resizebox{0.9\linewidth}{!}{
\begin{tabular}{l|cccc}
    \toprule
    \textbf{Model} & \textbf{Total Acc.} & \textbf{Normal Acc.} & \textbf{Anomaly Acc.} & \textbf{Time (s)} \\
    \midrule
    GPT-4o & 73.58 & 91.67 & 50.00 & 5.0058 \\
    Retrieval(unified) & 65.57 & 75.83 & 52.17 &  0.0711 \\
    Retrieval(separate) & 75.00 & 90.00 & 55.43 & 0.0586 \\
    Qwen2.5-VL-3B(ft) & 83.96 & 98.33 & 65.22 & 1.5496 \\
    Qwen2.5-VL-7B(ft) & 87.26 & 95.00 & 77.17 & 2.0107 \\
    \bottomrule
    COMMET(3B) & 84.43 & 98.33 & 66.30 & 1.3440 \\
    COMMET(7B) & 87.26 & 95.00 & 77.17 & 1.8117 \\
\end{tabular}
}
\label{tab:latency}
\end{table}

%% file: tex/5_conclusion.tex
\section{Conclusion}
\label{sec:conclusion}

We propose \emph{COMMET}, a system for human-induced conflicts in
mobile manipulation of everyday tasks. Then we demonstrate how we classify and collect potential real-world human-robot conflicts. Next, we present our COMMET's hybrid detection mechanism and its user preference-based solution prediction. In preliminary studies, we analyze the system's performance and latency, comparing it with GPT-4o. We also design a user-friendly interface and discuss an effective real-world deployment workflow to facilitate future user studies and real-world experiments.

%% file: tex/6_appendix.tex
\onecolumn

\section*{APPENDIX}

The appendix is structured as follows:

\begin{itemize}
\item Data Collection in Section~\ref{sup:data}.
\item Conflict Detection in Section~\ref{sup:detection}.
\item User Preference Prediction in Section~\ref{sup:preference}.
\item System Analysis~\ref{sup:exp}.
\end{itemize}

\section{Data Collection}
\label{sup:data}

\subsection{Conflict Type}
\textit{Human Interaction Conflict} is primarily caused by active human actions, while \textit{Human Occupancy Conflict} is typically results from passive human behaviors.  
\textit{Object State Conflict} are more hidden compared to other types, even when everything seems normal, changes in object states may hinder the execution of current step. For example, your target object is not available now, the door is closed so you can't enter the room, or the receptacle is already full. \textit{Goal Absence Conflict} is primarily determined by checking whether the objects in the image match with the requirements in current step.  
\textit{Human-Interaction Conflict} and \textit{Human Occupancy Conflict} are mainly identified based on the speech input and the image which contains humans. \textit{Object State Conflict} requires an overall analysis of the task, current step, and the state of the objects observed in the image.

\subsection{Collection Principle}

In the beginning, we assume the robot is performing a daily task and is in a certain stage, and design conflict scenarios that might occur in real life. Then a collector will simulate the robot's first-person perspective and record the current visual observation at the moment that conflict occurs. Other people will act as task-irrelevant disturbances.

Since the robot's detection is a continuous process, we also collect dynamic trajectory data. Similarly, a total task and each step are designed first, but the difference is that the collector needs to simulate the robot to complete the entire task and record a complete trajectory. This would include all detection data during normal execution, conflict occurrence, and conflict resolution.

After the collection is complete, we'll annotate the data. The dynamic trajectory will be saved as one frame every 0.5 seconds. Then all images are labeled with the final task, the current step, the background speech text, and the conflict type (Normal if there is no conflict). Furthermore, we augment the data by appropriately modifying these contents. Finally, the format of our data can be referred in Fig~\ref{fig:interface}.

\section{Conflict Detection}
\label{sup:detection}

In the retrieval stage, we only need to compare with the samples in the database. For model inference, our system instruction is as follows:
\begin{lstlisting}
system = """You are a household robot and now you are finishing the given task in the real world.
However, due to the dynamics of real-world environments and unpredictable activities of family members, the step you are currently executing may be affected or experience conflicts, making it impossible to continue execution.
Common conflicts can be categorized into the following four types:
1) Goal Absence Conflict: the robot is going to operate but the target is missing.
2) Human Interaction Conflict: the robot is executing tasks but another person want to interact with it.
3) Human Occupancy Conflict: Another person's activities hinders the plan execution or occupies the goal.
4) Object State Conflict: current goal state can't satisfy the task needs or current step can't continue due to object state changes. (e.g. the road is obstructed or the container is full)

Human Interaction Conflict is primarily caused by active human actions, while Human Occupancy Conflict is typically results from passive human behaviors.  
Object State Conflict are more hidden compared to other types, even when everything seems normal, changes in object states may hinder the execution of current step. For example, your target object is not available now, the door is closed so you can't enter the room, or the receptacle is already full.

I will provide the following information:  
1) Observation: A first-head view observation image of current scene.
1) Task: The final user task for you to complete.  
2) Current Step: The step you are currently executing.  
3) Speech: Detected background human speech.  

Goal Absence Conflict is primarily determined by checking whether the objects in the image match with the requirements in current step.  
Human Interaction Conflict and Human Occupancy Conflict are mainly identified based on the speech input and the image which contains humans.  
Object State Conflict requires an overall analysis of the task, current step, and the state of the objects observed in the image.

Note that Speech may either be irrelevant noise or a source of conflict. In the image, consider the distance between objects and yourself to determine if a conflict might occur.
You need to carefully analyze these inputs to determine whether any conflict exists in the current scene.
If a conflict is detected, output the type of conflict and the explicit reason of the conflict, otherwise output Normal.
"""
\end{lstlisting}

\section{User Preference Prediction}
\label{sup:preference}

We use GPT-4o to summarize user preferences and predict options. Our system instruction is as follows:
\begin{lstlisting}
descriptions = {
    "Goal Absence Conflict": "the robot is going to operate but the target is missing.",
    "Human Occupancy Conflict": "another person's activities occupy the specific space or object, thus hindering the task execution.",
    "Object State Conflict": "current goal state can't satisfy the task needs, or the current step can't continue due to object state changes (e.g. the door is closed or the container is full).",
    "Human Interaction Conflict": "the robot is executing tasks but another person attempts to command the robot or interact with it."
}

system = f"""You are a home robot currently executing a given task, but a {type} occurs now. This kind of conflict indicates that {description}.
Now you have following options to resolve the conflict: {solution}

All these options can solve the conflict, and the final decision should depend on the user's preferences. I will provide 5 user cases, each case includes an image of the conflict scenario, the executed task, the current step, detailed conflict information, and an emergency level.
The emergency level has three layers and indicates the severity of the conflict in the current context: the higher level means current situation is more urgent for the user, and the system should prioritize quickly resolving it. Conversely, a lower emergency level means it's acceptable for the robot to make certain compromises to resolve the conflict in a more harmonious way.
You need to learn the user's preferences in different situations from these cases, and choose the option that best aligns with the user's preference.
Note that the user's preferences may vary from different context. You need to learn more general patterns from the chosen options and emergency levels, and relate them to the current scenario in order to make a better decision.

Now I will provide you with 5 cases and the final scenario which you need to make a decision.
"""
\end{lstlisting}
The system instruction will vary form the current conflict type to achieve a more precise output.

\newpage

\section{System Analysis}
\label{sup:exp}

When selecting the optimal value for each parameter, we adhere to the following principles: First and foremost is the overall accuracy. If multiple values yield the same or very similar accuracy, the value with the higher anomaly accuracy is prioritized. If there are still multiple such values, the one with the smallest average inference time will be chosen.

\begin{figure*}[h]
\centering
\includegraphics[width=\linewidth]{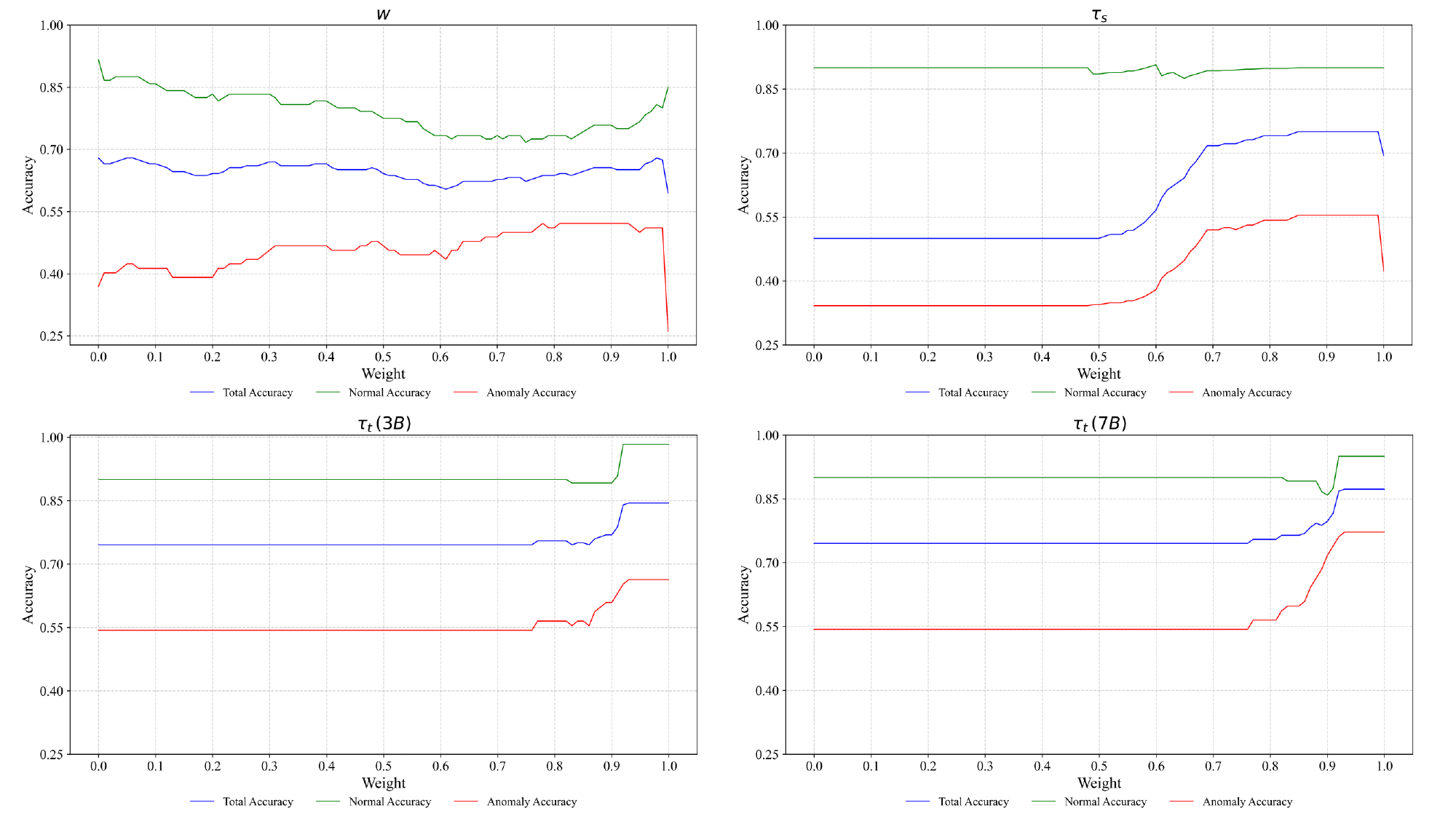}
\caption{Changes in accuracy corresponding to different parameter values.}
\label{fig:exp}
\end{figure*}

The changes in accuracy are shown in Fig~\ref{fig:exp}. As $w$ only affects the calculation of the final confidence score, it won't affect the average detection time. As for $\tau_s$, speech retrieval is less time-consuming than multi-modal retrieval and only occurs when the background sound exists, so its impact on the average detection time is also minimal. For $\tau_t$, due to the huge gap in time consumption between model inference and retrieval, the value of $\tau_t$ has a great impact on the average detection time. The detection time starts from a small value, increases sharply within a short interval, then changes slowly and eventually stabilizes.

